\documentclass{article}

\PassOptionsToPackage{numbers, compress}{natbib}
\usepackage[final]{nips_2016}

\usepackage[utf8]{inputenc} 
\usepackage[T1]{fontenc}    
\usepackage{hyperref}       
\usepackage{url}            
\usepackage{booktabs}       
\usepackage{amsfonts}       
\usepackage{nicefrac}       
\usepackage{microtype}      

\usepackage{algorithm}
\usepackage{algorithmic}
\usepackage{latexsym}
\usepackage{amsmath}
\usepackage{amsthm}
\usepackage{amssymb}
\usepackage{graphicx}
\usepackage{xspace}
\usepackage{tabularx}
\usepackage{multicol}
\usepackage{multirow}
\usepackage{wrapfig}
\usepackage{comment}
\usepackage{listings}
\usepackage{color,soul}

\newcommand{\bmx}[0]{\begin{bmatrix}}
\newcommand{\emx}[0]{\end{bmatrix}}

\newcommand{\vect}[1]{\mathbf{#1}}

\newcommand{\matr}[1]{\mathbf{#1}}

\newcommand{\vc}[0]{\vect{c}}
\newcommand{\ve}[0]{\vect{e}}

\newcommand{\vh}[0]{\vect{h}}
\newcommand{\vv}[0]{\vect{v}}

\newcommand{\mW}[0]{\matr{W}}

\newcommand{\G}[0]{\mathcal{G}}

\newcommand{\E}[0]{\mathcal{E}}
\newcommand{\N}[0]{\mathcal{N}}
\newcommand{\A}[0]{\mathcal{A}}
\newcommand{\T}[0]{\mathcal{T}}
\newcommand{\DD}[0]{\mathcal{D}}

\newcommand{\RR}[0]{\mathbb{R}}

\newcommand{\Exp}[0]{\mathbb{E}}

\DeclareMathOperator*{\argmax}{\arg \max}

\begin{document}

\title{End-to-End Goal-Driven Web Navigation}

\author{
  Rodrigo Nogueira \\
  Tandon School of Engineering\\
  New York University\\
  \texttt{rodrigonogueira@nyu.edu} \\
  \And
  Kyunghyun Cho \\
  Courant Institute of Mathematical Sciences \\
  New York University\\
  \texttt{kyunghyun.cho@nyu.edu} \\
}


\maketitle

\begin{abstract}
    We propose a goal-driven web navigation as a benchmark task for evaluating
    an agent with abilities to understand natural language and plan on partially
    observed environments. In this challenging task, an agent navigates through
    a website, which is represented as a graph consisting of web pages as nodes
    and hyperlinks as directed edges, to find a web page in which a query
    appears. The agent is required to have sophisticated high-level reasoning
    based on natural languages and efficient sequential decision-making
    capability to succeed. We release a software tool, called WebNav, that
    automatically transforms a website into this goal-driven web navigation
    task, and as an example, we make WikiNav, a dataset constructed from the
    English Wikipedia. We extensively evaluate different variants of neural net
    based artificial agents on WikiNav and observe that the proposed goal-driven
    web navigation well reflects the advances in models, making it a suitable
    benchmark for evaluating future progress. Furthermore, we extend the WikiNav
    with question-answer pairs from {\it Jeopardy!} and test the proposed agent
    based on recurrent neural networks against strong inverted index based
    search engines. The artificial agents trained on WikiNav outperforms the
    engined based approaches, demonstrating the capability of the proposed
    goal-driven navigation as a good proxy for measuring the progress in
    real-world tasks such as focused crawling and question-answering.
\end{abstract}

\section{Introduction}

In recent years, there have been many exciting advances in building an
artificial agent, which can be trained with one learning algorithm, to solve
many relatively large-scale, complicated tasks (see, e.g.,
\citep{mnih2015human,risi2014neuroevolution,koutnik2014evolving}.) In much of
these works, target tasks were computer games such as Atari games
\citep{mnih2015human} and racing car game \citep{koutnik2014evolving}.

These successes have stimulated researchers to apply a similar learning
mechanism to {\em language}-based tasks, such as multi-user dungeon (MUD) games
\citep{narasimhan2015language,he2015deep}. Instead of visual perception, an
agent perceives the state of the world by its written description. A set of
actions allowed to the agent is either fixed or dependent on the current state.
This type of task can efficiently evaluate the agent's ability of not only in
planning but also language understanding. 

We, however, notice that these MUD games do not exhibit the complex nature of
natural languages to the full extent. For instance, the largest game world
tested by \citet{narasimhan2015language} uses a vocabulary of only 1340 unique
words, and the largest game tested by \citet{he2015deep} uses only 2258 words.
Furthermore, the description of a state at each time step is almost always
limited to the visual description of the current scene, lacking any use of
higher-level concepts present in natural languages.

In this paper, we propose a goal-driven web navigation as a large-scale
alternative to the text-based games for evaluating artificial agents with
natural language understanding and planning capability. The proposed goal-driven
web navigation consists of the whole website as a graph, in which the web pages
are nodes and hyperlinks are directed edges. An agent is given a query, which
consists of one or more sentences taken from a randomly selected web page in the
graph, and navigates the network, starting from a predefined starting node, to
find a target node in which the query appears. Unlike the text-based games, this
task utilizes the existing text as it is, resulting in a large vocabulary with a
truly natural language description of the state. Furthermore, the task is more
challenging as the action space greatly changes with respect to the state in
which the agent is. 

We release a software tool, called WebNav, that converts a given website into a
goal-driven web navigation task. As an example of its use, we provide WikiNav,
which was built from English Wikipedia. We design artificial agents based on
neural networks (called NeuAgents) trained with supervised learning, and report
their respective performances on the benchmark task as well as the performance
of human volunteers. We observe that the difficulty of a task generated by
WebNav is well controlled by two control parameters; (1) the maximum number of
hops from a starting to a target node $N_h$ and (2) the length of query $N_q$.

Furthermore, we extend the WikiNav with an additional set of queries that are
constructed from {\it Jeopardy!} questions, to which we refer by
WikiNav-Jeopardy. We evaluate the proposed NeuAgents against the three
search-based strategies; (1) SimpleSearch, (2) Apache Lucene and (3) Google
Search API. The result in terms of document recall indicates that the NeuAgents
outperform those search-based strategies, implying a potential for the proposed
task as a good proxy for practical applications such as question-answering and
focused crawling.

\section{Goal-driven Web Navigation}
\label{sec:task}

A task $\T$ of goal-driven web navigation is characterized by 
\begin{align}
    \label{eq:task}
    \T=(\A, s_S, \G, q, R, \Omega).
\end{align}

The world in which an agent $\A$ navigates is represented as a graph
$\G=(\N,\E)$.  The graph consists of a set of nodes $\N=\left\{ s_i
\right\}_{i=1}^{N_{\N}}$ and a set of directed edges $\E=\left\{e_{i,j}
\right\}$ connecting those nodes.  Each node represents a page of the website,
which, in turn, is represented by the natural language text $\DD(s_i)$ in it.
There exists an edge going from a page $s_i$ to $s_j$ if and only if there is a
hyperlink in $\DD(s_i)$ that points to $s_j$.  One of the nodes is designated as
a starting node $s_S$ from which any navigation begins. A target node is the one
whose natural language description contains a query $q$, and there may be more
than one target node.

At each time step, the agent $\A$ {\em reads} the natural language description
$\DD(s_t)$ of the current node in which the agent has landed. At no point, the
whole world, consisting of the nodes and edges, nor its structure or map (graph
structure without any natural language description) is visible to the agent, thus
making this task {\em partially observed}.

Once the agent $\A$ reads the description $\DD(s_i)$ of the current node $s_i$,
it can take one of the actions available. A set of possible actions is defined
as a union of all the outgoing edges $e_{i,\cdot}$ and the {\em stop} action,
thus making the agent have {\em state-dependent} action space.

Each edge $e_{i,k}$ corresponds to the agent jumping to a next node $s_k$, while
the stop action corresponds to the agent declaring that the current node $s_i$
is one of the target nodes.  Each edge $e_{i,k}$ is represented by the
description of the next node $\DD(s_k)$. In other words, deciding which
action to take is equivalent to taking a peek at each neighboring node and
seeing whether that node is likely to lead ultimately to a target node. 

The agent $\A$ receives a reward $R(s_i, q)$ when it chooses the stop action.
This task uses a simple binary reward, where 
\begin{align*}
    R(s_i, q) = \left\{ \begin{array}{l l}
            1,&\text{if }q \subseteq \DD(s_i) \\
            0,&\text{otherwise}
        \end{array}\right.
\end{align*}

\paragraph{Constraints}

It is clear that there exists an ultimate policy for the agent to succeed at
every trial, which is to traverse the graph breadth-first until the agent finds
a node in which the query appears. To avoid this kind of degenerate policies,
the task includes a set of four rules/constraints $\Omega$:
\begin{enumerate}
    \itemsep -.2em
    \item An agent can follow at most $N_n$ edges at each node.
    \item An agent has a finite memory of size smaller than $\T$.
    \item An agent moves up to $N_h$ hops away from $s_S$.
    \item A query of size $N_q$ comes from at least two hops away from the
        starting node.
\end{enumerate}

The first constraint alone prevents degenerate policies, such as breadth-first
search, forcing the agent to make good decisions as possible at each node. The
second one further constraints ensure that the agent does not cheat by using
earlier trials to reconstruct the whole graph structure (during test time) or to
store the entire world in its memory (during training.) The third constraint,
which is optional, is there for computational consideration. The fourth
constraint is included because the agent is allowed to read the content of a
next node.

\section{WebNav: Software}

As a part of this work, we build and release a software tool which turns a
website into a goal-driven web navigation task.\footnote{
    The source code and datasets are publicly available at
    \url{https://drive.google.com/folderview?id=0B5LbsF7OcHjqUFhWQ242bzdlTWc&usp=sharing}.
}
We call this tool {\em WebNav}.  Given a starting URL, the WebNav reads the
whole website, constructs a graph with the web pages in the website as nodes.
Each node is assigned a unique identifier $s_i$. The text content of each node
$\DD(s_i)$ is a cleaned version of the actual HTML content of the corresponding
web page.  The WebNav turns intra-site hyperlinks into a set of edges $e_{i,j}$. 

In addition to transforming a website into a graph $\G$ from
Eq.~\eqref{eq:task}, the WebNav automatically selects queries from the nodes'
texts and divides them into training, validation, and test sets.  We ensure that
there is no overlap among three sets by making each target node, from which a
query is selected, belongs to only one of them.

Each generated example is defined as a tuple
\begin{align}
    \label{eq:example1}
    X = (q, s^*, p^*)
\end{align}
where $q$ is a query from a web page $s^*$, which was found following a randomly
selected path $p^*=(s_S, \ldots, s^*)$. In other words, the WebNav starts from a
starting page $s_S$, random-walks the graph for a predefined number of steps
($N_h/2$, in our case), reaches a target node $s^*$ and selects a query $q$ from
$\DD(s^{*})$.  A query consists of $N_q$ sentences and is selected among the
top-5 candidates in the target node with the highest average TF-IDF, thus
discouraging the WebNav from choosing a trivial query.

For the evaluation purpose alone, it is enough to use only a query $q$ itself as
an example. However, we include both one target node (among potentially many
other target nodes) and one path from the starting node to this target node
(again, among many possible connecting paths) so that they can be exploited when
training an agent. They are not to be used when evaluating a trained agent.

\begin{table}[t]
    \centering
    \caption{Dataset Statistics of WikiNav-4-*,
    WikiNav-8-*, WikiNav-16-* and WikiNav-Jeopardy.}
    \label{tab:stat}

    \begin{tabular}{r c c c c}
        & {\bf WikiNav-4-*} & {\bf WikiNav-8-*} & {\bf WikiNav-16-*} & {\bf WikiNav-Jeopardy}\\ 
        \toprule
        {\bf Train} & 6.0k & 1M & 12M & 113k \\
        \midrule
        {\bf Valid} & 1k & 20k & 20k & 10k \\
        \midrule
        {\bf Test } & 1k & 20k & 20k & 10k \\
    \end{tabular}

    \vspace{-4mm}
\end{table}

\section{WikiNav: A Benchmark Task}

With the WebNav, we built a benchmark goal-driven navigation task using
Wikipedia as a target website. We used the dump file of the English Wikipedia
from September 2015, which consists of more than five million web pages. We
built a set of separate tasks with different levels of difficulty by varying the
maximum number of allowed hops $N_h \in \left\{ 4, 8, 16 \right\}$ and the size
of query $N_q \in \left\{1, 2, 4 \right\}$. We refer to each task by
\mbox{WikiNav-$N_h$-$N_q$}. 

For each task, we generate training, validation and test examples from the pages
half as many hops away from a starting page as the maximum number of hops
allowed.\footnote{
    This limit is an artificial limit we chose for computational reasons.
} 
We use ``Category:Main topic classifications'' as a starting node $s_S$.

As a minimal cleanup procedure, we excluded meta articles whose titles start
with ``Wikipedia''. Any hyperlink that leads to a web page outside Wikipedia is
removed in advance together with the following sections: ``References'',
``External Links'', ``Bibliography'' and ``Partial Bibliography''.

\begin{wraptable}{r}{0.4\textwidth}
    \centering
    \begin{tabular}{r || c | c }
        & Hyperlinks & Words \\
        \hline
        \hline
        Avg. & 4.29 & 462.5 \\
        $\sqrt{\text{Var}}$ & 13.85 & 990.2 \\
        Max & 300 & 132881 \\
        Min & 0 & 1 \\
    \end{tabular}
    \caption{Per-page statistics of English Wikipedia.}
    \label{tab:wiki_stat}

    \vspace{-10mm}
\end{wraptable}

In Table~\ref{tab:wiki_stat}, we present basic per-article statistics of the
English Wikipedia. It is evident from these statistics that the world of
\mbox{WikiNav-$N_h$-$N_q$} is large and complicated, even after the cleanup
procedure. 

We ended up with a fairly small dataset for \mbox{WikiNav-4-*}, but large for
\mbox{WikiNav-8-*} and \mbox{WikiNav-16-*}. See Table~\ref{tab:stat} for
details.

\subsection{Related Work: Wikispeedia}

This work is indeed not the first to notice the possibility of a website, or
possibly the whole web, as a world in which intelligent agents explore to
achieve a certain goal. One most relevant recent work to ours is perhaps
Wikispeedia from \citep{west2009wikispeedia,west2012automatic,west2012human}.

West~et~al.~\citep{west2009wikispeedia,west2012automatic,west2012human} proposed
the following game, called {\it Wikispeedia}. The game's world is nearly
identical to the goal-driven navigation task proposed in this work. More
specifically, they converted ``Wikipedia for Schools''
, which contains approximately 4,000 articles as of 2008, into a graph whose
nodes are articles and directed edges are hyperlinks.  From this graph, a pair
of nodes is randomly selected and provided to an agent. 

The agent's goal is to start from the first node, navigate the graph and reach
the second node. Similarly to the WikiNav, the agent has access to the text
content of the current nodes and all the immediate neighboring nodes. One major
difference is that the target is given as a whole article, meaning that there is
a single target node in the Wikispeedia while there may be multiple target
nodes in the proposed WikiNav.

From this description, we see that the goal-driven web navigation is a
generalization and re-framing of the Wikispeedia. First, we constrain a query to
contain less information, making it much more difficult for an agent to navigate
to a target node.
Furthermore, a major research question by \citet{west2012human} was to ``{\it 
    understand how humans navigate and find the information they are looking for
},'' whereas in this work we are fully focused on proposing an automatic tool
to build a challenging goal-driven tasks for designing and evaluating {\em
artificial} intelligent agents.

\section{WikiNav-Jeopardy: {\it Jeopardy!} on WikiNav}

One of the potential practical applications utilizing the goal-drive navigation is
question-answering based on world knowledge. In this Q\&A task, a query is a
question, and an agent navigates a given information network, e.g., website, to
retrieve an answer. In this section, we propose and describe an extension of the
WikiNav, in which query-target pairs are constructed from actual {\it Jeopardy!}
question-answer pairs.  We refer to this extension of WikiNav by {\it
WikiNav-Jeopardy}. 

\begin{table}[tb]
    \centering
    \caption{Sample query-answer pairs from WikiNav-Jeopardy.}
    \label{tab:JeoQuestions}
    \begin{tabular}{ll}
        {\bf Query} & {\bf Answer} \\ 
        \toprule
        \small For the last 8 years of his life, Galileo was under &
        \multirow{2}{*}{Copernicus} \\
        \small house arrest for espousing this man's theory. & \\
        \midrule
        \small In the winter of 1971-72, a record 1,122 inches of snow fell &
        \multirow{2}{*}{Washington} \\
        \small at Rainier Paradise Ranger Station in this state. & \\
        \midrule
        \small This company's Accutron watch, introduced in 1960, &
        \multirow{2}{*}{Bulova} \\
        \small had a guarantee of accuracy to within one minute a month. & 
    \end{tabular}
    \vspace{-4mm}
\end{table}

We first extract all the question-answer pairs from {\it J! Archive}\footnote{
    \url{http://www.j-archive.com}
}, which has more than 300k such pairs. We keep only those pairs whose answers
are titles of Wikipedia articles, leaving us with 133k pairs. We divide those
pairs into 113k training, 10k validation, and 10k test examples while carefully
ensuring that no article appears in more than one partition. Additionally, we do
not shuffle the original pairs to ensure that the train and test examples are
from different episodes.

For each training pair, we find one path from the starting node ``Main Topic
Classification'' to the target node and include it for supervised learning. For
reference, the average number of hops to the target node is 5.8, the standard
deviation is 1.2, and the maximum and minimum are 2 and 10, respectively.
See Table~\ref{tab:JeoQuestions} for sample query-answer pairs.

%
%

\section{NeuAgent: Neural Network based Agent}
\label{sec:reagent}

\subsection{Model Description}

\paragraph{Core Function} 

The core of the NeuAgent is a parametric function $f_{\text{core}}$ that takes
as input the content of the current node $\phi_c (s_i)$ and a query $\phi_q
(q)$, and that returns the hidden state of the agent. This parametric function
$f_{\text{core}}$ can be implemented either as a feedforward neural network
$f_{\text{ff}}$:
\begin{align*}
    \vh_t = f_{\text{ff}}(\phi_c (s_i), \phi_q (q))
\end{align*}
which does not take into account the previous hidden state of the agent or as a
recurrent neural network $f_{\text{rec}}$:
\begin{align*}
    \vh_t = f_{\text{rec}}(\vh_{t-1}, \phi_c(s_i), \phi_q (q)).
\end{align*}
We refer to these two types of agents by {\it NeuAgent-FF} and {\it
NeuAgent-Rec}, respectively. For the NeuAgent-FF, we use a single $\tanh$
layer, while we use long short-term memory (LSTM) units
\citep{hochreiter1997long}, which have recently become {\em de facto} standard,
for the NeuAgent-Rec.

\begin{wrapfigure}{i}{0.4\textwidth}
    \centering
    \includegraphics[width=\linewidth]{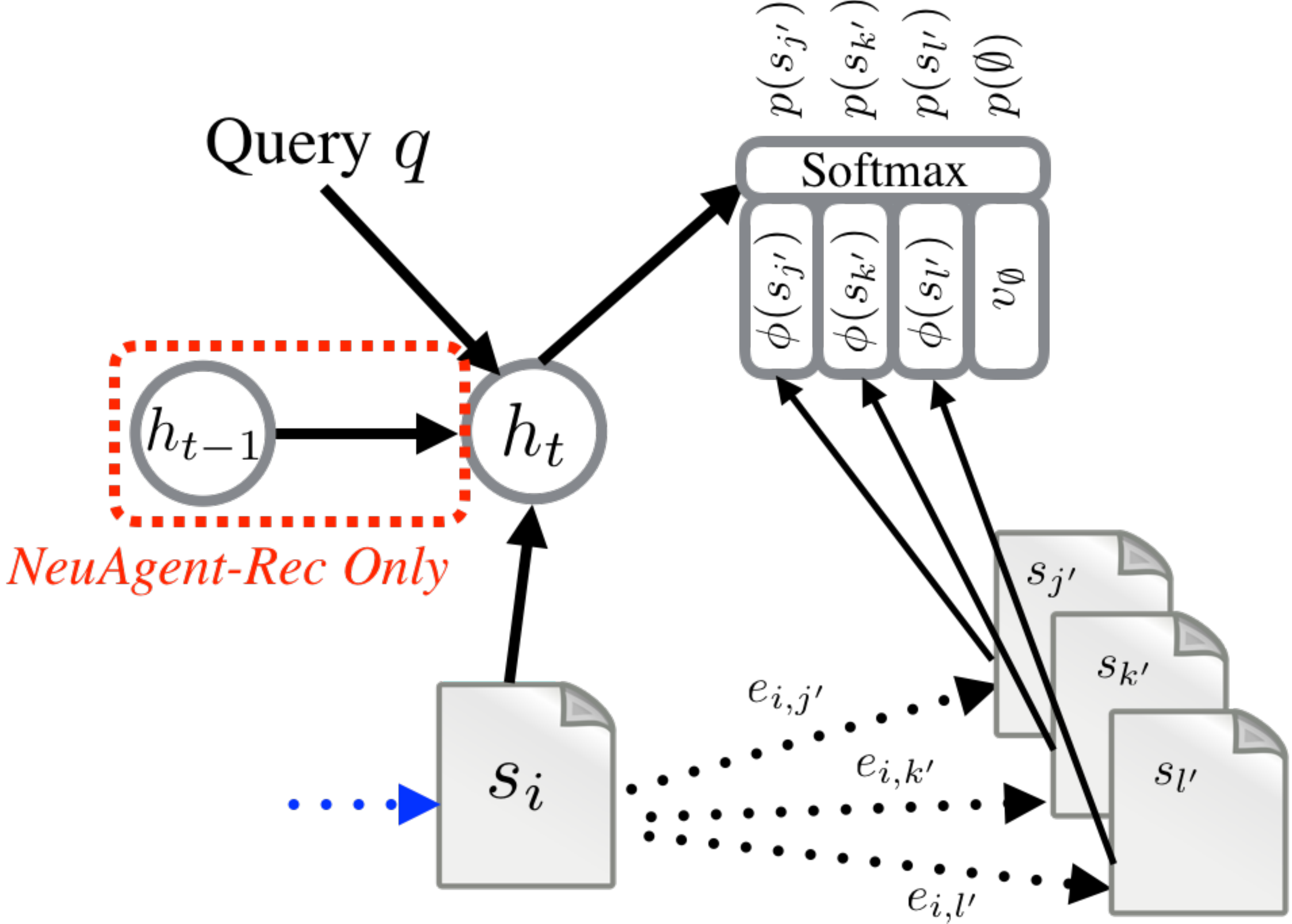}

    \vspace{-3mm}
    \caption{Graphical illustration of a single step performed by the baseline
    model, NeuAgent.}
    \label{fig:agent}

    \vspace{8mm}
\end{wrapfigure}

Based on the new hidden state $\vh_t$, the NeuAgent computes the probability
distribution over all the outgoing edges $e_i$. The probability of each outgoing
edge is proportional to the similarity between the hidden state $\vh_t$ such
that
\begin{align}
    \label{eq:edge_p}
    p(e_{i, j}|\tilde{p}) \propto \exp \left( \phi_c(s_j)^\top \vh_t \right).
\end{align}
Note that the NeuAgent peeks at the content of the next node $s_j$ by
considering its vector representation $\phi_c(s_j)$.  In addition to all the
outgoing edges, we also allow the agent to {\em stop} with the probability
\begin{align}
    \label{eq:stop_p}
    p(\emptyset|\tilde{p}) \propto \exp \left( \vv_{\emptyset}^\top \vh_t \right),
\end{align}
where the stop action vector $\vv_{\emptyset}$ is a trainable parameter. In the
case of NeuAgent-Rec, all these (unnormalized) probabilities are conditioned on
the history $\tilde{p}$ which is a sequence of actions (nodes) selected by the
agent so far. We apply a {\em softmax} normalization on the unnormalized
probabilities to obtain the probability distribution over all the possible
actions at the current node $s_i$.

The NeuAgent then selects its next action based on this action probability
distribution (Eqs.~\eqref{eq:edge_p} and \eqref{eq:stop_p}). If the stop action
is chosen, the NeuAgent returns the current node as an answer and receives a
reward $R(s_i, q)$, which is one if correct and zero otherwise. If the agent
selects one of the outgoing edges, it moves to the selected node and repeats
this process of {\em reading} and {\em acting}. 

See Fig.~\ref{fig:agent} for a single step of the described NeuAgent.

\paragraph{Content Representation}

The NeuAgent represents the content of a node $s_i$ as a vector $\phi_c(s_i) \in
\RR^d$. In this work, we use a continuous bag-of-words vector for each document:
\[
    \phi_c(s_i) = \frac{1}{|\DD(s_i)|} \sum_{k=1}^{|\DD(s_i)|} \ve_k.
\]
Each word vector $\ve_k$ is from a pretrained continuous bag-of-words model
\citep{mikolov2013efficient}. These word vectors are fixed throughout training. 

\paragraph{Query Representation}

In the case of a query, we consider two types of representation. The first one
is a continuous bag-of-words (BoW) vector, just as used for representing the
content of a node. The other one is a dynamic representation based on the
attention mechanism~\citep{bahdanau2014neural}. 

In the attention-based query representation, the query is first projected into a
set of context vectors. The context vector of the $k$-th query word is
\[
    \vc_k = \sum_{k'=k-u/2}^{k+u/2} \mW_{k'} \ve_{k'},
\]
where $\mW_{k'} \in \RR^{d \times d}$ and $\ve_{k'}$ are respectively a
trainable weight matrix and a pretrained word vector. $u$ is the window size.
Each context vector is scored at each time step $t$ by $\beta_k^t =
f_{\text{att}}(\vh_{t-1}, \vc_k)$ w.r.t. the previous hidden state of the
NeuAgent, and all the scores are normalized to be positive and sum to one, i.e.,
$\alpha_k^t = \frac{\exp(\beta_k^t)}{\sum_{l=1}^{|q|} \exp(\beta_l^t)}$. These
normalized scores are used as the coefficients in computing the weighted-sum of
query words to result in a query representation at time $t$: 
\[
    \phi_q(q) = \frac{1}{|q|} \sum_{k=1}^{|q|} \alpha_k^t \vc_k.
\]
Later, we empirically compare these two query representations.


\subsection{Inference: Beam Search}
\label{sec:beamsearch}

Once the NeuAgent is trained, there are a number of approaches to using it for
solving the proposed task. The most naive approach is simply to let the agent
make a greedy decision at each time step, i.e., following the outgoing edge with
the highest probability $\argmax_{k} \log p(e_{i,k}|\ldots)$. A better approach
is to exploit the fact that the agent is allowed to explore up to $N_n$ outgoing
edges per node. We use a simple, forward-only beam search with the beam width
capped at $N_n$. The beam search simply keeps the $N_n$ most likely traces, in
terms of $\log p(e_{i, k} |\ldots)$, at each time step.

\subsection{Training: Supervised Learning}

In this paper, we investigate supervised learning, where we train the agent to
follow an example trace $p^*=(s_S, \ldots, s^*)$ included in the training set at
each step (see Eq.~\eqref{eq:example1}). In this case, the cost per training
example is
\begin{align}
    \label{eq:c_sup}
    C_{\text{sup}} = -\log p(\emptyset|p^*, q) - \sum_{k=1}^{|p^*|} \log p(p^*_k |
    p^*_{<k}, q).
\end{align}
This per-example training cost is fully differentiable with respect to all the
parameters of the neural network, and we use stochastic gradient descent (SGD)
algorithm to minimize this cost over the whole training set, where the gradients
can be computed by backpropagation \citep{rumelhart1986learning}. This allows the
entire model to be trained in an end-to-end fashion, in which the query-to-target
performance is optimized directly.


\section{Human Evaluation}

One unique aspect of the proposed task is that it is very difficult for an
average person who was not trained specifically for finding information by
navigating through an information network. There are a number of reasons behind
this difficulty. First, the person must be familiar with, via training, the
graph structure of the network, and this often requires many months, if not
years, of training. Second, the person must have in-depth knowledge of a broad
range of topics in order to make a connection via different concepts between the
themes and topics of a query to a target node. Third, each trial requires the
person carefully to read the whole content of the nodes as she navigates, which
is a time-consuming and exhausting job.


We asked five volunteers to try up to 20 four-sentence-long queries\footnote{
    In a preliminary study with other volunteers, we found that, when the
    queries were shorter than $4$, they were not able to solve enough trials for
    us to have meaningful statistics.
    }
randomly selected from the test sets of \mbox{WikiNav-$\left\{ 4, 8, 16
\right\}$-4} datasets. They were given up to two hours, and they were allowed to
choose up to the same maximum number of explored edges per node $N_n$ as the
NeuAgents (that is, $N_n=4$), and also were given the option to give up. The
average reward was computed as the fraction of correct trials over all the
queries presented.

\begin{table}[t]
    \centering
    \caption{The average reward by the NeuAgents and humans on the
        test sets of \mbox{WikiNav-$N_h$-$N_q$}.
    }
    \label{tab:results}
    \small
    \begin{tabular}{c ccc||ccc|ccc|ccc}
        & & & & \multicolumn{3}{c|}{$N_q=1$} & \multicolumn{3}{c|}{2} & \multicolumn{3}{c}{4} \\
        \hline
        & $f_{\text{core}}$ & {\scriptsize \# Layers $\times$ \# Units} & $\phi_q$ & $N_h=4$ & 8 & 16 & 4 & 8 & 16 & 4 & 8 & 16 \\
        \toprule
        (a) & {\normalsize $f_{\text{ff}}$} & $1\times512$ {\scriptsize $\tanh$} & BoW & 21.5 & 4.7 & 1.2 & 40.0 & 9.2 & 1.9 & 45.1 & 12.9 & 2.9 \\
        (b) & {\normalsize $f_{\text{rec}}$} & $1\times512$ {\scriptsize LSTM} &  BoW & 22.0 & 5.1 & 1.7 & 41.1 & 9.2 & 2.1 & 44.8 & 13.3 & 3.6 \\
        (c) & {\normalsize $f_{\text{rec}}$} & $8\times2048$ {\scriptsize LSTM} &  BoW & 17.7 & 10.9 & 8.0 & 35.8 & 19.9 & 13.9 & 39.5 & 28.1 & 21.9 \\
        (d) & {\bf \normalsize $f_{\text{rec}}$} & {\bf $8\times2048$
    {\scriptsize LSTM}} &  {\bf Att} & \textbf{22.9} & \textbf{15.8} & \textbf{12.5} & \textbf{41.7} & \textbf{24.5} & \textbf{17.8} & \textbf{46.8} & \textbf{34.2} & \textbf{28.2} \\
        (e) & \multicolumn{3}{c||}{Humans} & - & - & - & - & - & - & 14.5 & 8.8 & 5.0 
    \end{tabular}
    \vspace{-4mm}
\end{table}

\section{Results and Analysis}
\label{sec:result}

\subsection{WikiNav}

We report in Table~\ref{tab:results} the performance of the NeuAgent-FF and
NeuAgent-Rec models on the test set of all nine \mbox{WikiNav-$\left\{ 4, 8,
16\right\}$-$\left\{ 1, 2, 4\right\}$} datasets. In addition to the proposed
NeuAgents, we also report the results of the human evaluation.

We clearly observe that the level of difficulty is indeed negatively correlated
with the query length $N_q$ but is positively correlated with the maximum number
of allowed hops $N_h$. The latter may be considered trivial, as the size of the
search space grows exponentially with respect to $N_h$, but the former is not.
The former negative correlation confirms that it is indeed easier to solve the
task with more information in a query. We conjecture that the agent requires
more in-depth understanding of natural languages and planning to overcome the
lack of information in the query to find a path toward a target node.

The NeuAgent-FF and NeuAgent-Rec shares similar performance when the maximum
number of allowed hops is small ($N_h=4$), but NeuAgent-Rec ((a) vs. (b))
performs consistently better for higher $N_h$, which indicates that having
access to history helps in long-term planning tasks. We also observe that the
larger and deeper NeuAgent-Rec ((b) vs (c)) significantly outperforms the
smaller one, when a target node is further away from the starting node $s_S$.

\begin{figure}[b]
    \centering
    \includegraphics[width=\columnwidth,clip=True,trim=30 0 0 0]{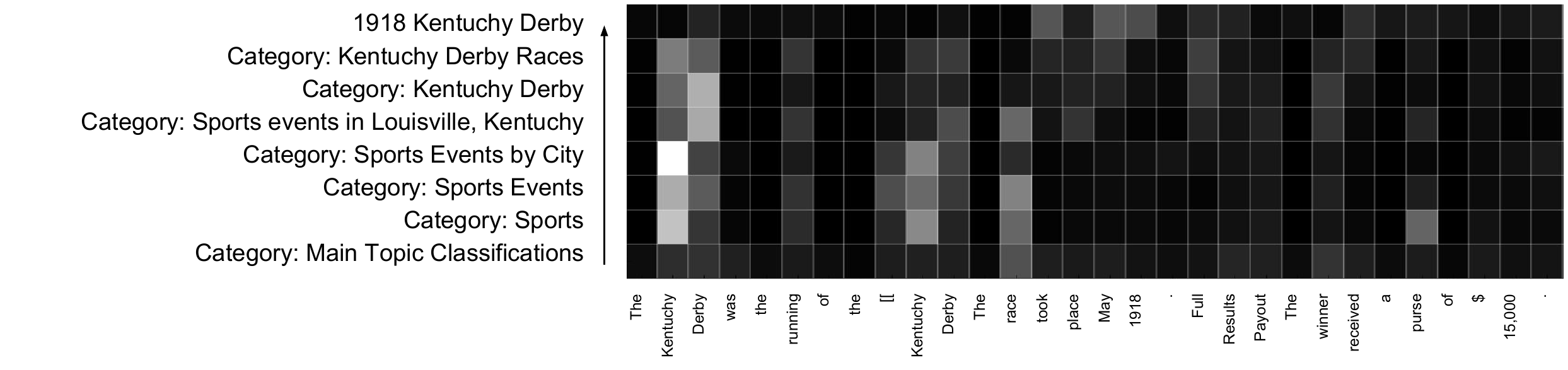}
    \vspace{-4mm}
    \caption{Visualization of the attention weights over a test query. The
    horizontal axis corresponds to the query words, and the vertical axis to the
    articles titles visited.}
    \label{fig:query_att_vis}
\end{figure}

The best performing model in (d) used the attention-based query representation,
especially as the difficulty of the task increased ($N_q \downarrow$ and $N_h
\uparrow$), which supports our claim that the proposed task of goal-driven web
navigation is a challenging benchmark for evaluating future progress. In
Fig.~\ref{fig:query_att_vis}, we present an example of how the attention weights
over the query words dynamically evolve as the model navigates toward a target
node.

%
%

The human participants generally performed worse than the NeuAgents. We
attribute this to a number of reasons. First, the NeuAgents are trained
specifically on the target domain (Wikipedia), while the human participants have
not been. Second, we observed that the volunteers were rapidly exhausted from
reading multiple articles in sequence. In other words, we find the proposed
benchmark, WebNav, as a good benchmark for machine intelligence but not for
comparing it against human intelligence.

\subsection{WikiNav-Jeopardy}

\paragraph{Settings} 
We test the best model from the previous experiment (NeuAgent-Rec with 8 layers
of 2048 LSTM units and the attention-based query representation) on the
WikiNav-Jeopardy. We evaluate two training strategies. The first strategy is
straightforward supervise learning, in which we train a NeuAgent-Rec on
WikiNav-Jeopardy from scratch. In the other strategy, we pretrain a NeuAgent-Rec
first on the WikiNav-16-4 and finetune it on WikiNav-Jeopardy.

We compare the proposed NeuAgent against three search strategies. The first one,
{\it SimpleSearch}, is a simple inverted index based strategy. SimpleSearch
scores each Wikipedia article by the TF-IDF weighted sum of words that co-occur
in the articles and a query and returns top-$K$ articles. Second, we use Lucene,
a popular open source information retrieval library, in its default
configuration on the whole Wikipedia dump. Lastly, we use Google Search
API\footnote{
    \url{https://cse.google.com/cse/}
}, while restricting the domain to \url{wikipedia.org}.

Each system is evaluated by document recall at $K$ (Recall@$K$). We vary $K$ to
be 1, 4 or 40. In the case of the NeuAgent, we run beam search with width set to
$K$ and returns all the $K$ final nodes to compute the document recall. 

\begin{table}[htb]
    \centering
    \caption{Recall on WikiNav-Jeopardy. {\scriptsize $(\star)$ Pretrained on
    WikiNav-16-4.}}
    \label{tab:jeoresults}

    \small
    \begin{tabular}{l c ||c|c|c}
        Model & Pre$^\star$ & Recall@1 & Recall@4 & Recall@40 \\ 
        \toprule
        NeuAgent & & 13.9 & 20.2 & 33.2 \\ 
        NeuAgent & $\checkmark $& \textbf{18.9} & \textbf{23.6} & \textbf{38.3} \\ 
        \midrule
        \multicolumn{2}{l||}{SimpleSearch} & 5.4 & 12.6 & 28.4 \\ 
        \multicolumn{2}{l||}{Lucene} & 6.3 & 14.7 & 36.3 \\ 
        \multicolumn{2}{l||}{Google} & 14.0 & 22.1 & 25.9 \\ 
    \end{tabular}
\end{table}

\paragraph{Result and Analysis}

In Table~\ref{tab:jeoresults}, we report the results on WikiNav-Jeopardy. The
proposed NeuAgent clearly outperforms all the three search-based strategies,
when it was pretrained on the WikiNav-16-4. The superiority of the pretrained
NeuAgent is more apparent when the number of candidate documents is constrained
to be small, implying that the NeuAgent is able to accurately rank a correct
target article.  Although the NeuAgent performs comparably to the other
search-based strategy even without pretraining, the benefit of pretraining on
the much larger WikiNav is clear.

We emphasize that these search-based strategies have
access to all the nodes for each input query. The NeuAgent, on the other hand,
only observes the nodes as it visits during navigation. This success clearly
demonstrates a potential in using the proposed NeuAgent pretrained with a
dataset compiled by the proposed WebNav for the task of focused
crawling~\citep{chakrabarti1999focused,alvarez2007deepbot}, which is an
interesting problem on its own, as much of the content available on the Internet
is either hidden or dynamically generated~\citep{alvarez2007deepbot}.



\section{Conclusion}
\label{sec:conclusion}

In this work, we describe a large-scale goal-driven web navigation task and argue
that it serves as a useful test bed for evaluating the capabilities of
artificial agents on natural language understanding and planning. We release
a software tool, called WebNav, that compiles a given website into a goal-driven
web navigation task. As an example, we construct WikiNav from Wikipedia using
WebNav. We extend WikiNav with {\it Jeopardy!} questions, thus creating
WikiNav-Jeopardy.
We evaluate various neural net based agents on WikiNav and WikiNav-Jeopardy.
Our results show that more sophisticated agents have better performance,
thus supporting our claim that this task is well suited to evaluate future
progress in natural language understanding and planning. Furthermore, we show
that our agent pretrained on WikiNav
outperforms two strong inverted-index based search engines on the
WikiNav-Jeopardy. These empirical results support our claim on the usefulness
of the proposed task and agents in challenging applications such as focused
crawling and question-answering.

\section*{Acknowledgments}
KC thanks the support by Facebook, Google (Google Faculty Award 2016) and NVidia (GPU Center
of Excellence 2015-2016).
RN is funded by Coordenação de Aperfeicoamento de Pessoal de Nível Superior (CAPES).

\appendix

\vskip 0.2in

\bibliography{webnav}

\begin{thebibliography}{14}
\providecommand{\natexlab}[1]{#1}
\providecommand{\url}[1]{\texttt{#1}}
\expandafter\ifx\csname urlstyle\endcsname\relax
  \providecommand{\doi}[1]{doi: #1}\else
  \providecommand{\doi}{doi: \begingroup \urlstyle{rm}\Url}\fi

\bibitem[{\'A}lvarez et~al.(2007){\'A}lvarez, Raposo, Pan, Cacheda, Bellas, and
  Carneiro]{alvarez2007deepbot}
Manuel {\'A}lvarez, Juan Raposo, Alberto Pan, Fidel Cacheda, Fernando Bellas,
  and V{\'\i}ctor Carneiro.
\newblock Deepbot: a focused crawler for accessing hidden web content.
\newblock In \emph{Proceedings of the 3rd international workshop on Data
  enginering issues in E-commerce and services: In conjunction with ACM
  Conference on Electronic Commerce (EC'07)}, pages 18--25. ACM, 2007.

\bibitem[Bahdanau et~al.(2014)Bahdanau, Cho, and Bengio]{bahdanau2014neural}
Dzmitry Bahdanau, Kyunghyun Cho, and Yoshua Bengio.
\newblock Neural machine translation by jointly learning to align and
  translate.
\newblock In \emph{ICLR 2015}, 2014.

\bibitem[Chakrabarti et~al.(1999)Chakrabarti, Van~den Berg, and
  Dom]{chakrabarti1999focused}
Soumen Chakrabarti, Martin Van~den Berg, and Byron Dom.
\newblock Focused crawling: a new approach to topic-specific web resource
  discovery.
\newblock \emph{Computer Networks}, 31\penalty0 (11):\penalty0 1623--1640,
  1999.

\bibitem[He et~al.(2015)He, Chen, He, Gao, Li, Deng, and Ostendorf]{he2015deep}
Ji~He, Jianshu Chen, Xiaodong He, Jianfeng Gao, Lihong Li, Li~Deng, and Mari
  Ostendorf.
\newblock Deep reinforcement learning with an unbounded action space.
\newblock \emph{arXiv preprint arXiv:1511.04636}, 2015.

\bibitem[Hochreiter and Schmidhuber(1997)]{hochreiter1997long}
Sepp Hochreiter and J{\"u}rgen Schmidhuber.
\newblock Long short-term memory.
\newblock \emph{Neural computation}, 9\penalty0 (8):\penalty0 1735--1780, 1997.

\bibitem[Koutn{\'\i}k et~al.(2014)Koutn{\'\i}k, Schmidhuber, and
  Gomez]{koutnik2014evolving}
Jan Koutn{\'\i}k, J{\"u}rgen Schmidhuber, and Faustino Gomez.
\newblock Evolving deep unsupervised convolutional networks for vision-based
  reinforcement learning.
\newblock In \emph{Proceedings of the 2014 conference on Genetic and
  evolutionary computation}, pages 541--548. ACM, 2014.

\bibitem[Mikolov et~al.(2013)Mikolov, Chen, Corrado, and
  Dean]{mikolov2013efficient}
Tomas Mikolov, Kai Chen, Greg Corrado, and Jeffrey Dean.
\newblock Efficient estimation of word representations in vector space.
\newblock \emph{arXiv preprint arXiv:1301.3781}, 2013.

\bibitem[Mnih et~al.(2015)Mnih, Kavukcuoglu, Silver, Rusu, Veness, Bellemare,
  Graves, Riedmiller, Fidjeland, Ostrovski, et~al.]{mnih2015human}
Volodymyr Mnih, Koray Kavukcuoglu, David Silver, Andrei~A Rusu, Joel Veness,
  Marc~G Bellemare, Alex Graves, Martin Riedmiller, Andreas~K Fidjeland, Georg
  Ostrovski, et~al.
\newblock Human-level control through deep reinforcement learning.
\newblock \emph{Nature}, 518\penalty0 (7540):\penalty0 529--533, 2015.

\bibitem[Narasimhan et~al.(2015)Narasimhan, Kulkarni, and
  Barzilay]{narasimhan2015language}
Karthik Narasimhan, Tejas Kulkarni, and Regina Barzilay.
\newblock Language understanding for text-based games using deep reinforcement
  learning.
\newblock \emph{arXiv preprint arXiv:1506.08941}, 2015.

\bibitem[Risi and Togelius(2014)]{risi2014neuroevolution}
Sebastian Risi and Julian Togelius.
\newblock Neuroevolution in games: State of the art and open challenges.
\newblock \emph{arXiv preprint arXiv:1410.7326}, 2014.

\bibitem[Rumelhart et~al.(1986)Rumelhart, Hinton, and
  Williams]{rumelhart1986learning}
David Rumelhart, Geoffrey Hinton, and Ronald Williams.
\newblock Learning representations by back-propagating errors.
\newblock \emph{Nature}, pages 323--533, 1986.

\bibitem[West and Leskovec(2012{\natexlab{a}})]{west2012automatic}
Robert West and Jure Leskovec.
\newblock Automatic versus human navigation in information networks.
\newblock In \emph{ICWSM}, 2012{\natexlab{a}}.

\bibitem[West and Leskovec(2012{\natexlab{b}})]{west2012human}
Robert West and Jure Leskovec.
\newblock Human wayfinding in information networks.
\newblock In \emph{21st International World Wide Web Conference}, pages
  619--628. ACM, 2012{\natexlab{b}}.

\bibitem[West et~al.(2009)West, Pineau, and Precup]{west2009wikispeedia}
Robert West, Joelle Pineau, and Doina Precup.
\newblock Wikispeedia: An online game for inferring semantic distances between
  concepts.
\newblock In \emph{IJCAI}, pages 1598--1603, 2009.

\end{thebibliography}
\bibliographystyle{plainnat}

\end{document}